\documentclass{article}
\usepackage{subcaption}

\usepackage[nonatbib,final]{neurips_2020}

\usepackage[utf8]{inputenc} 
\usepackage[T1]{fontenc}    
\usepackage{hyperref}       
\hypersetup{
    colorlinks=true,
    linkcolor=blue,
    filecolor=magenta,      
    urlcolor=cyan,
}
\usepackage{url}            
\usepackage{booktabs}       
\usepackage{amsfonts}       
\usepackage{nicefrac}       
\usepackage{microtype}      
\usepackage{acronym}
\usepackage{graphicx}
\usepackage{xcolor}
\usepackage{siunitx}        
\usepackage{enumitem} 

\acrodef{dl}[DL]{Deep Learning}
\acrodef{ml}[ML]{Machine Learning}
\acrodef{ndwi}[NDWI]{Normalized Difference Water Index}

\acrodef{lidar}[LiDAR]{Light Detection and Ranging}
\acrodef{dem}[DEM]{Digital Elevation Model}
\acrodef{nir}[NIR]{Near-Infrared}
\acrodef{nhd}[NHD]{National Hydrography Dataset}
\acrodef{hag}[HAG]{High Above Ground}
\acrodef{wv3}[WV3]{WorldView 3}
\acrodef{aois}[AoIs]{Areas of Interest}
\acrodef{aoi}[AoI]{Area of Interest}
\acrodef{VHR}[VHR]{very-high resolution}
\acrodef{3DEP}[3DEP]{3D Elevation Program}

\title{{
Pix2Streams: Dynamic Hydrology Maps from Satellite-LiDAR Fusion
}}

\author{
   Dolores Garcia\thanks{Equal contribution (listed in alphabetical order).}~
   \thanks{\texttt{dolores.garcia@imdea.org; gonzalo.mateo-garcia@uv.es; freddie@fdl.ai}}\\
  Frontier Development Lab \\
  \And
  Gonzalo Mateo-Garcia\footnotemark[1] ~\footnotemark[2]{}\\
  University of Valencia\\
  \AND
  Hannes Bernhardt \\
   Arizona State University\\
  \And
  Ron Hagensieker \\
  osir.io\\
  \And
  Ignacio G. Lopez-Francos\\
  NASA Ames Research Center  \\
  \AND
  Jonathan Stock\\
  USGS  \\
   \And
  Guy Schumann\\
  RSS-Hydro \\
  University of Bristol\\
  \And
 Kevin Dobbs\\
  Frontier Development Lab\\
  \AND
  Freddie Kalaitzis\footnotemark[1] ~\footnotemark[2]{}\\
  University of Oxford\\Frontier Development Lab
}

\begin{document}

\maketitle
\begin{abstract}
Where are the Earth's streams flowing right now?
Inland surface waters expand with floods and contract with droughts, so there is no one map of our streams.
Current satellite approaches are limited to monthly observations that map only the widest streams.
These are fed by smaller tributaries that make up much of the dendritic surface network but whose flow is unobserved.
A complete map of our daily waters can give us an early warning for where droughts are born: the receding tips of the flowing network.
Mapping them over years can give us a map of impermanence of our waters, showing where to expect water, and where not to.
To that end,
we feed the latest high-res sensor data to multiple deep learning models in order to map these flowing networks every day, stacking the times series maps over many years.
Specifically,
\textbf{i)} we enhance water segmentation to $50$ cm/pixel resolution, a 60$\times$ improvement over previous state-of-the-art results. Our U-Net trained on 30-40cm WorldView3 images can detect streams as narrow as 1-3m (30-60$\times$ over SOTA).
Our multi-sensor, multi-res variant, \textbf{WasserNetz}, fuses a multi-day window of 3m PlanetScope imagery with 1m LiDAR data, to detect streams 5-7m wide.
Both U-Nets produce a water probability map at the pixel-level.
\textbf{ii)} We integrate this water map over a DEM-derived synthetic valley network map to produce a snapshot of flow at the stream level.
\textbf{iii)} We apply this pipeline, which we call \textbf{Pix2Streams}, to a 2-year daily PlanetScope time-series of three watersheds in the US to produce the first high-fidelity dynamic map of stream flow frequency.
The end result is a new map that, if applied at the national scale, could fundamentally improve how we manage our water resources around the world.
\end{abstract}

\section{Introduction}
\vspace{-1ex}

The United Nations 2020 Sustainable Development Goals report \cite{sustainable_2020} states that \textit{"water scarcity affects more than 40\% of the global population and is projected to increase. Over 1.7 billion people are currently living in river basins where water use exceeds recharge"}.
Small and medium sized streams make up a significant portion of the flowing network.
The receding tips of this network could serve as an early warning for where droughts are born~\cite{steinemann2006developing, hannaford2011examining, svensson2017statistical}.
In the United States, the \ac{nhd}~\cite{geological_survey_2019} contains large rivers, some smaller streams, as well as lakes that have been mapped during aerial and on-ground mapping campaigns since the early 20th century. A subset of these streams are monitored by gages that continually record the water levels at discrete locations.
However, small and intermittent streams (widths < 90 m) are inadequately mapped and not monitored, even though they contribute 48\% of the total surface water extent of the stream network globally~\cite{allen2018global}.

Open data from satellite missions, like Sentinel and Landsat, have been used for water mapping applications \cite{mueller_water_2016,pekel_high-resolution_2016,isikdogan_rivamap_2017,yang_monthly_2020}.
Sentinel and Landsat offer temporally sparse images, in 5 and 16 day intervals, respectively, and at resolutions (10-30m/px) that are too low for detecting small and intermittent streams.
At the opposite extreme, \ac{VHR} imagery, like from Maxar's WorldView satellites, can be processed into very high resolution (<5m) maps over smaller areas.
Such satellites are generally task-based, that is, they acquire images on customer demand, so multi-temporal coverage is not guaranteed.
As a trade-off, the PlanetScope constellation has high-res (3m), daily, wall-to-wall coverage, offering a \ac{VHR}-grade imagery that is temporally dynamic and spatially extensive.



\begin{figure}
    \centering
    \includegraphics[width=.1995\linewidth]{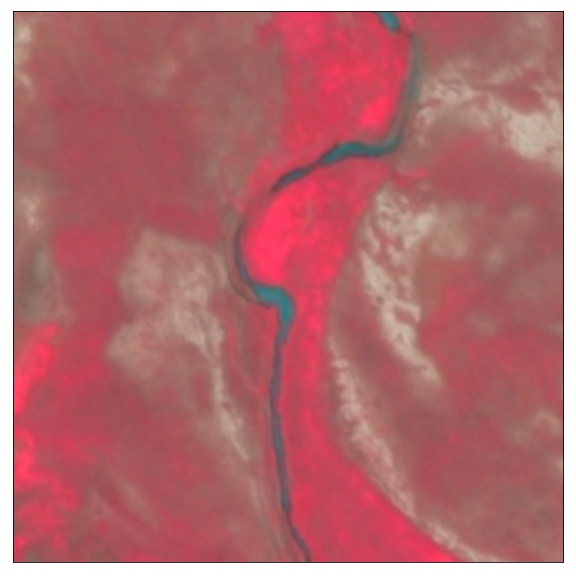}\hspace*{-.1cm}
    \includegraphics[width=.1995\linewidth]{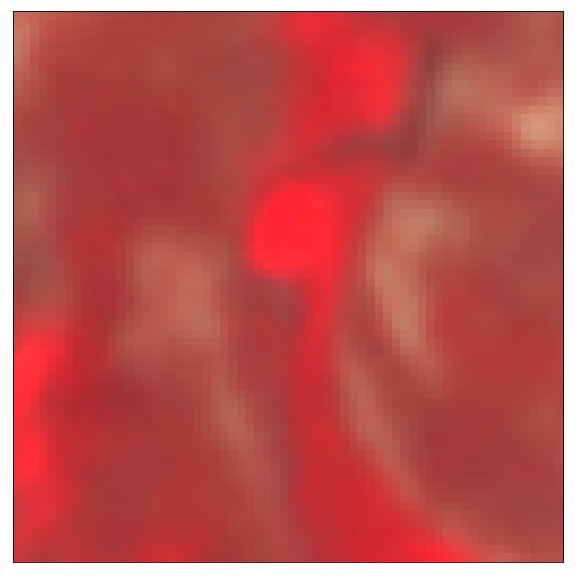}\hspace*{-.1cm}
    \includegraphics[width=.194\linewidth]{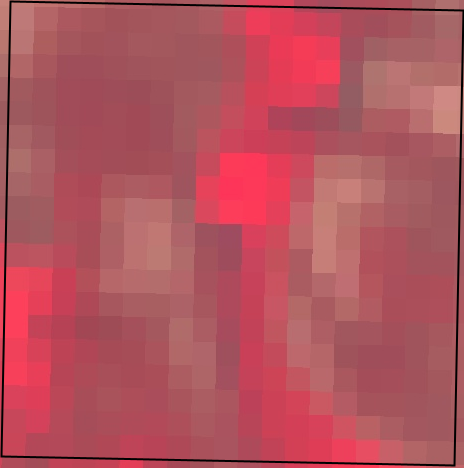}\hspace*{-.1cm}
    \includegraphics[width=.1995\linewidth]{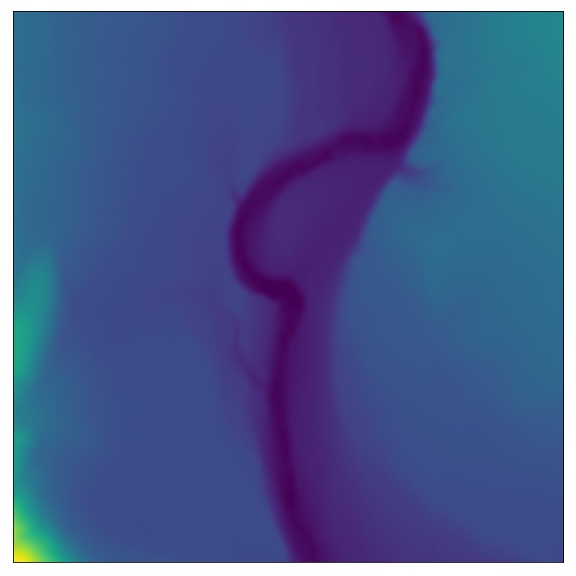}\hspace*{-.1cm}
    \includegraphics[width=.1995\linewidth]{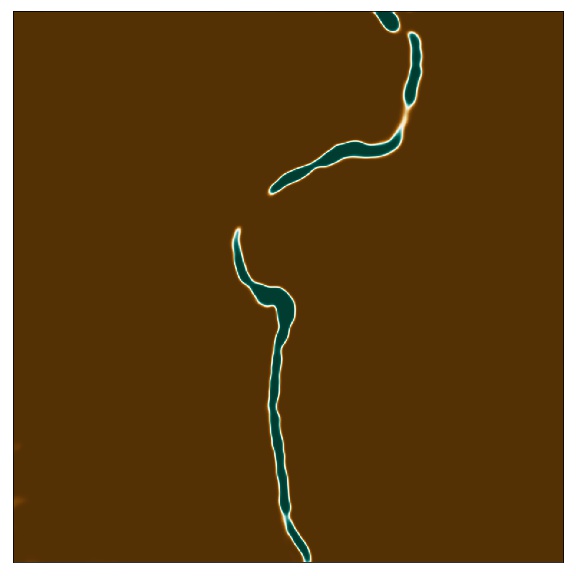}
    \caption{\small Narrow (5-2m) river in the South Dakota AoI. \textbf{Left to right:} \ac{wv3} (50cm/px), PlanetScope (3m/px), Sentinel-2 (30m/px), \ac{dem} (1m/px), water probability map (50cm/px) of the same area. \ac{wv3}, All satellite images are acquired on the same day, and are shown with a false RGB composite of \ac{nir}, Red, and Green bands.}
    \label{fig:resolutions}
\end{figure}

\paragraph{From pixels to streams}\footnote{Video on the steps from the pixel-level map, to the \textbf{stream-level map}: \href{http://bit.ly/pix2streams}{bit.ly/pix2streams} 
}
We show that, by combining \ac{VHR} with high cadence, we can monitor previously unseen streams at near real-time (daily).
Our \textbf{Pix2streams} pipeline consists of two functional primitives:
\begin{enumerate}[leftmargin=*]
    \item {\bf Pixel-level mapping:} A U-Net (SOTA semantic segmentation model~\cite{ronneberger2015u}) creates pixel-level water probability maps.
    Our variant, WasserNetz, fuses multiple sensors at different resolutions and outputs a water probability map at the resolution of the ground-truth annotation layer (0.5m/px).
    \item {\bf Stream-level mapping:} We aggregate these probabilities over a synthetic valley network, derived from 1m LiDAR-based digital elevation maps (\ac{dem}), to obtain a stream-level hydrology map.
\end{enumerate}

We apply this pipeline to a 2-year daily time-series of 3m PlanetScope images, in three US watersheds, to obtain a stream-level \textbf{dynamic hydrology map}, that is, a collection of $N$ time-series (for $N$ streams) of the stream's surface \% covered by water, daily.
Through this map, we can now monitor stream flow frequency, and therefore quantify water availability on a daily, monthly, or yearly basis.

\paragraph{Contributions}
\begin{enumerate}[label=\Alph*, leftmargin=*]
    \vspace{-1ex}
    \item We develop WasserNetz, a {\bf water segmentation model for PlanetScope imagery} trained on a custom dataset of manually labeled polygons, derived from same-day very high-resolution \ac{wv3} images (0.5m/px).
    We show that water detection accuracy is improved by fusing the PlanetScope imagery with USGS high-resolution \ac{lidar} derived products.
    \item We analyze and compare the performance of WasserNetz {\bf with models trained on higher resolution (30-40cm/px) \ac{wv3} images} on a manually annotated test dataset.
    Also, we study the sensitivity of the models \textbf{w.r.t. the stream width}.
    \item We produce a {\bf fundamentally new hydrology map at the stream level}, by fusing the pixel-level output of WasserNetz with the synthetic valley network, derived from the LiDAR-based \ac{dem}.
\end{enumerate}

\section{Related Work}
\vspace{-1ex}

Most of the work on mapping and monitoring of rivers uses medium-resolution (30m/px) imagery from Landsat or Sentinel-2 \cite{mueller_water_2016,isikdogan_rivamap_2017,yang_monthly_2020}, and only recently has \ac{VHR} (sub-meter) imagery been used, albeit with a broad focus on all water bodies~\cite{randall2019geographic,feng2018water}.
Landsat imagery was used to derive continental and global maps of large river systems, and to estimate river width and surface area at the global scale~\cite{allen2018global,allen2015patterns}.
Through Google's Earth Engine, Landsat was also used to conduct a multi-annual global inventory of inland waters~\cite{pekel_high-resolution_2016}, enabling monthly estimations of water permanence.
Limited by Landsat's \SI{30}{\metre/px} resolution, both studies can reliably detect only streams wider than $60-90$m (2-3 pixels).
Until today, this spatial limitation has been endemic in the detection of narrower streams~\cite{frasson2019global}.

Some aspects of our work are not directly comparable to these studies:
on the data side, due to the \textbf{unique spatial and temporal resolution} of our imagery and LiDAR derivatives.
On the application side, our pipeline for \textbf{daily \ac{VHR} river mapping} is the first of its kind.
On the algorithmic side, the U-Net~\cite{ronneberger2015u}, with its concurrent encoder-decoder bottlenecks of different resolution inputs, learns multi-scale features to ultimately inform a fine-grained segmentation.
This simple and intuitive idea is accepted as \textbf{a staple of semantic segmentation} in remote sensing~\cite{zhang2018road,wei2019multi,yao2018pixel,feng2018water,wang2020weakly}.

Instead of comparing architectures, our ablations in Section~\ref{sec:results} focus on the resolving capacity of similar U-Nets w.r.t. two fronts: i) training on 30-40cm \ac{wv3} vs. 3m PlanetScope + 1m LiDAR derivatives (WasserNetz); ii) stream width.

\section{Data} \label{sec:Data}
\vspace{-1ex}

We study three \ac{aois} across the western United States.
With limited resources for expert labeling, these \ac{aois} are chosen to provide a representative sample of the morphology, geology and vegetation of terrains.
Fig~\ref{fig:resolutions} shows an example of the three types of data used per AoI:

\textbf{\ac{wv3}:} a single mosaic acquired over multiple dates, with 30-50cm ground sample distance (GSD), 8 spectral bands (coastal blue to near-infrared (NIR)), atmospherically corrected and pansharpened.

\textbf{PlanetScope:} 2 years (April 2018 - April 2020) of daily snapshots, with 3m GSD, four bands (rgb + NIR).
We used an experimental version, corrected for the presence of clouds, radiometrically harmonized to Landsat, Sentinel-2 and MODIS information~\cite{houborg2018cubesat}.

\textbf{\ac{lidar} derivatives:} With the TauDEM suite~\cite{taudem_2015} we produce hydrology products from the  bare-earth 1m resolution \ac{3DEP} \ac{dem}, like the synthetic valley network, reach catchments, Strahler stream order, and a stream buffer (the distance from a given pixel to the nearest reach).
See \ref{sec:lidarder} for details on the derivatives used for training.


\textbf{Manually labeled polygons.} To train the model in a supervised manner, a set of georeferenced delineated polygons corresponding to `wet' and `dry' classes is annotated. These annotations focus on narrow streams and contain labels with shadows for both wet and dry, which are difficult to classify. 
Polygons are drawn using the \ac{wv3} images as a reference to exploit its higher resolution. The dataset contains a total of 5664 labels with 46.39$\%$ water polygons, distributed on the three \ac{aois}. We split the dataset into train (5121) and test sets (543), so that they do not overlap spatially.

\section{Methodology}

    \vspace{-1ex}
    \paragraph{Semantic segmentation model}
    We consider the architecture U-Net~\cite{ronneberger2015u} as a starting point.
    We modify it to have two downstream branches (PlanetScope images and \ac{lidar} products) each taking a different size input. The branches are merged when the size of the feature maps is the same. To provide the output at the resolution of the ground-truth (50cm/px) we add an upsampling convolution step. We refer to this network as \textbf{WasserNetz}. See Appendix \ref{sec:WasserNetz} for an illustration.

    \vspace{-1ex}
    \paragraph{Loss function}
    We use the binary cross entropy loss,  back-propagating only the pixels inside the rasterized polygon, where ground-truth is available. This is required since only pixels within those polygons are annotated. Also, the loss contribution of each example is weighted by the number of labeled pixels in its ground truth, i.e polygon-level loss. This seeks to mitigate the bias due to polygon size.
    
\begin{figure*}[ht]
    \centering
    \includegraphics[width=0.74\linewidth]{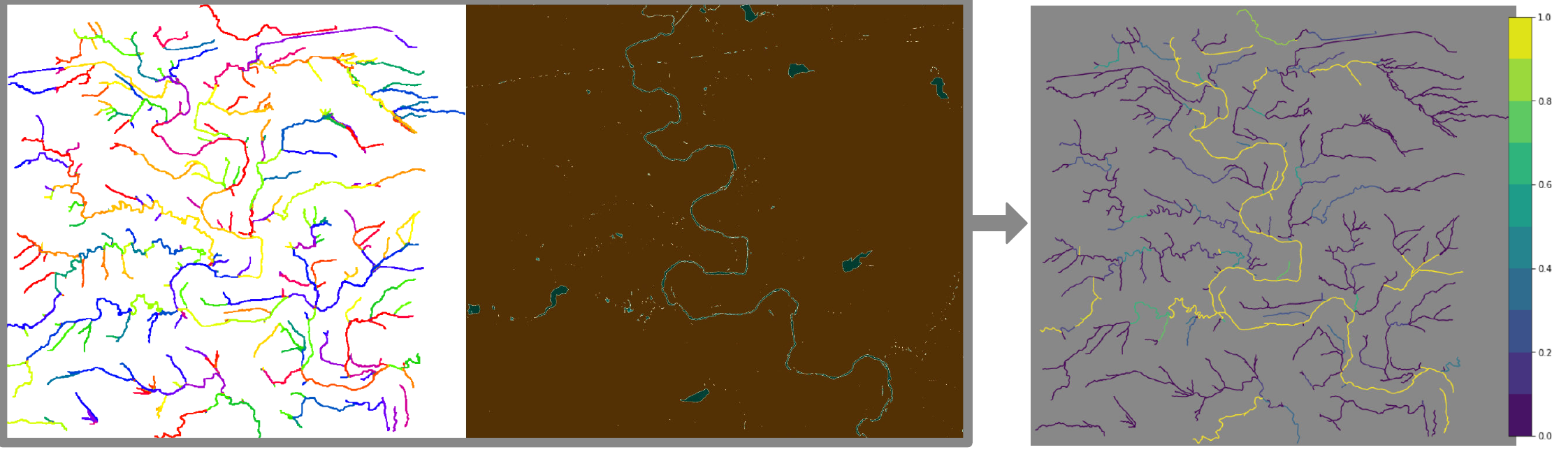} 
    \includegraphics[width=0.245\linewidth]{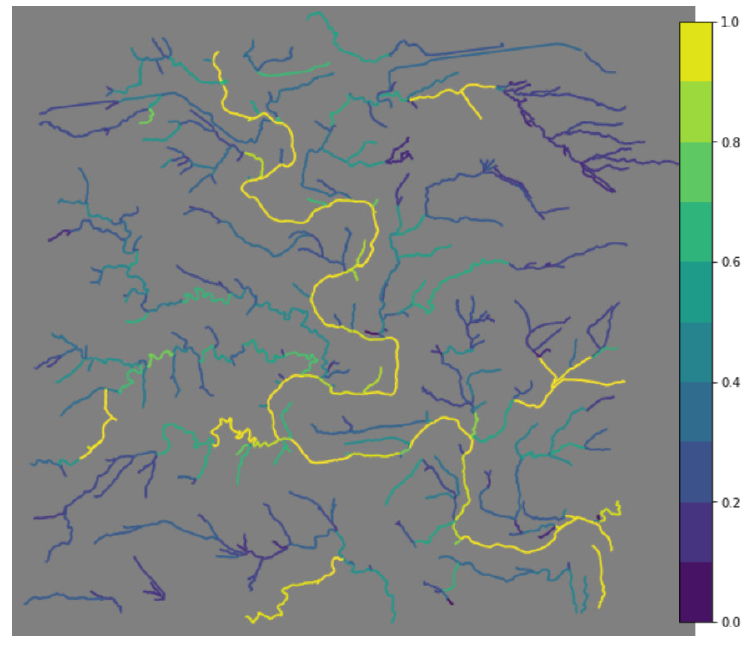}
    \caption{\small \textbf{Left to right:} 1) Synthetic valley network where each reach in the network is shown with a different color. 2) Pixel-level water probability map (colormap: [0-"dry"/ 1-"water"] $\rightarrow$ [brown, blue]) 3) derived stream-level hydrology map for a single day. 4) Annual stream-level hydrology map produced for 2019.}
    \label{figurestatic}
\end{figure*}

    \vspace{-1ex}
    \paragraph{Dynamic Hydrology Map}
    The derived synthetic valley network is a vector layer that delineates all the reaches where water can flow. For each reach we consider its catchment area, that is, the hydrological area unit in which all water ends up in the corresponding reach. Then we aggregate the model prediction (pixel-level water probability map) over each catchment area, to give the probability of having water. For an example, see Figure \ref{figurestatic}. This reach-level probability is computed by weighting the pixel-level probability map by the distance to the reach, and it represents the likelihood of finding water at any part of that reach. The reaches and their associated water probabilities give a static snapshot of flowing water in the stream network. Applying this process to a time-series of images we obtain the near-real time stream network\footnote{Video on steps from pixel-level map, to stream-level, to \textbf{dynamic hydrology map}: \href{http://bit.ly/pix2streams}{bit.ly/pix2streams}}.

\section{Results} \label{sec:results}
\vspace{-1ex}
We measure the performance of our water segmentation model with the accuracy at the pixel-level and polygon-level. The latter seek to balance wet and dry classes since dry polygons are usually larger. PR-curves,included in the appendix~\ref{sec:prcurves}, show similar behaviors. 
Table (\ref{tab:metrics}) shows the results for different model configurations. 
The multi-sensor segmentation (\ac{lidar}+PlanetScope) outperforms by $\textrm{3}\%$ the model trained only on Planet data. This shows that \ac{lidar}-derived products complement 3m imagery yielding a significant improvement.
The U-Net trained on \ac{wv3} (0.5m) outperforms WasserNetz by $\textrm{7}\%$. This is expected, since below a critical width some streams can be resolved at 0.5m while not at 3m resolution. The \ac{ndwi} approach~\cite{McFeeters1996}, shown as a baseline, shows the worst performance, with a 30-40$\%$ loss in accuracy over all models.

\begin{figure}
    \centering
    \includegraphics[width=1\linewidth]{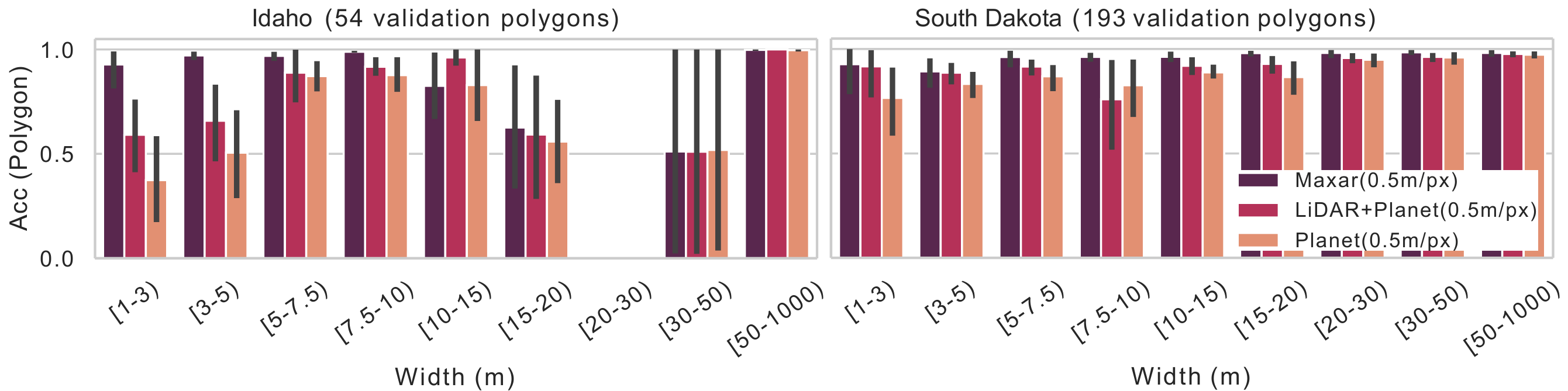}
    \vspace*{-5mm}
    \caption{\small Models performance on wet polygons of the test dataset stratified by stream width.}
    \label{annotatedpolygons}
    \vspace{-2mm}
\end{figure}

Also, we quantify the sensitivity of our pixel-level hydrology maps w.r.t. the width of the streams (Fig. \ref{annotatedpolygons}). We perform this analysis on a subset (280 polygons) of our test set that include width annotations enclosing the opposing banks of a water body.
A positive relationship between width and performance is observed for all models.
In South Dakota, the \ac{wv3}-trained model is the most robust but without a significant performance margin.
Also, South Dakota's landscape conditions seem to be favorable for WasserNetz (\ac{lidar}+PlanetScope), as they allow it to be competitive to the \ac{wv3}-trained U-Net.
Idaho (ID), with extensive shadows and canopy cover, poses an unfavorable environment for PlanetScope-trained models, whose performance degrades significantly on streams narrower than 5m. However, the performance gain of a multi-sensor model over a Planet only model is $18\%$ for narrower streams (<5m). This
suggests that our approach can lead to a large improvement in complex landscapes, especially for narrow streams (<5m).

Fig.~\ref{figurestatic} shows the annual (2019) stream-level water probabilities, produced by averaging daily stream-level probabilities, derived from PlanetScope images through our proposed pipeline.

\begin{table}[ht]
\begin{center}
    \vspace{-2mm}
    \footnotesize
    \caption{\small Polygon-level, pixel-level accuracy on test data. Mean (stdev) across 10 runs of different random seeds.}
    \vspace{1mm}
    \label{tab:metrics}
    \begin{tabular}{llllll}
    \toprule
    \textbf{Data} &\textbf{Model} &\textbf{Input m/px} &\textbf{Output m/px} & \textbf{Polygon-level} & \textbf{Pixel-level} \\
    \midrule
    \ac{lidar}+PlanetScope & WasserNetz & 3 / 1 & 0.5 &  88.24 (0.36) & 94.34 (0.43)  \\ 
    \ac{wv3}  & U-Net & 0.5 & 0.5 & 95.26 (0.23) & 96.87 (0.21) \\ 
    PlanetScope & U-Net &3 & 0.5 & 85.42 (0.24) & 91.60 (0.52)\\ 
    \ac{wv3} & NDWI  & 0.5 &  0.5 & 56.96 & 71.25 \\ 
    \bottomrule
\end{tabular}
\vspace{-5mm}
\end{center}
\end{table}

\section{Conclusions}
\vspace{-1ex}
Our multi-sensor deep-learning model can segment water at unprecedented scales, from a fusion of multispectral satellite imagery and \ac{lidar}-derived products, processed at their native resolutions. Fusing with LiDAR, compared to an optical-only model, improves the detection of narrow streams (< 5m) by $18\%$ in difficult areas (with shadows and canopy cover), and by $8\%$ in optically cooperative areas. We derive the first \textbf{dynamic hydrology map} from the predictions of the semantic segmentation model (WasserNetz) over a time-series of 2 years.
We emphasize the trade-off between spatial and temporal resolution in existing VHR instruments: even though this work pushes the observation frontier down to the scale of 1-3m stream widths (30-60$\times$ over SOTA) with WorldView3 data, our proposed \textbf{pix2streams} pipeline, and the main idea of our paper, rests on the availability of daily VHR imagery, as exemplified by the PlanetScope constellation.

\paragraph{Future work}
We plan to extend our results to other regions that are optically cooperative
and open-source a larger dataset of annotated polygons.
To further ease the interpretation of the stream-level probabilities in our dynamic hydrology maps, more work must be done to disentangle the spatial proportion of water coverage in a stream, from the uncertainty in a model prediction. The latter can be further broken down into \textit{aleatoric} uncertainty (noisy data and annotations) and \textit{epistemic} uncertainty (eg. ambiguity in features of shadow vs. water). Finally, we believe that the connectedness in dendritic stream networks, as an inductive bias, can inspire new architectures and losses that improve the data-efficiency and robustness in stream mapping.

\begin{ack}
This research was conducted during the 2020 Frontier Development Lab (FDL 2020) research accelerator, a public-private partnership with NASA in the USA and ESA in Europe. The authors gratefully acknowledge support from the United States Geological Survey (USGS), Google Cloud, NVIDIA, Planet. We thank the reviewers for their kind and constructive feedback, Yarin Gal for helpful comments on our manuscript, and all the people who supported us through FDL 2020 - alphabetically, Alan Rea, Alison Lowndes, Andrew Annex, Apoorva Shastry, Belina Raffy, Brad Neuberg, Dave Selkowitz, Giovanni Marchisio, Jack Eggleston, James Parr, Jason Stoker, Jodie Hughes, Julie Kiang, Kelsey Doerksen, Leo Silverberg, Massimo Mascaro, Milad Memarzadeh, Oleg Alexandrov, Rachel Sleeter, Richard Strange, Rob Emanuele, Roland Viger, Ross Beyer, Sara Jennnings, Scott Penberthy, Supratik Mukhopadhyay, Tyler Erickson, Yarin Gal, Zhe Jiang.
Gonzalo Mateo-Garcia has been partially supported by the Spanish Ministry of Science, Innovation and Universities under the projects TEC2016-77741-R and PID2019-109026RB-I00 (ERDF).
\end{ack}

\bibliographystyle{unsrt}
\bibliography{main.bib}

\begin{thebibliography}{10}

\bibitem{sustainable_2020}
United Nations~Statistics Division.
\newblock {\em The Sustainable Development Goals Report 2020}.
\newblock United Nations, 2020.

\bibitem{steinemann2006developing}
Anne~C Steinemann and Luiz~FN Cavalcanti.
\newblock Developing multiple indicators and triggers for drought plans.
\newblock {\em Journal of Water Resources Planning and Management},
  132(3):164--174, 2006.

\bibitem{hannaford2011examining}
Jamie Hannaford, Benjamin Lloyd-Hughes, Caroline Keef, Simon Parry, and
  Christel Prudhomme.
\newblock Examining the large-scale spatial coherence of european drought using
  regional indicators of precipitation and streamflow deficit.
\newblock {\em Hydrological Processes}, 25(7):1146--1162, 2011.

\bibitem{svensson2017statistical}
Cecilia Svensson, Jamie Hannaford, and Ilaria Prosdocimi.
\newblock Statistical distributions for monthly aggregations of precipitation
  and streamflow in drought indicator applications.
\newblock {\em Water Resources Research}, 53(2):999--1018, 2017.

\bibitem{geological_survey_2019}
U.S. Geological~Survey.
\newblock National hydrography dataset, 2016.

\bibitem{allen2018global}
George~H Allen and Tamlin~M Pavelsky.
\newblock Global extent of rivers and streams.
\newblock {\em Science}, 361(6402):585--588, 2018.

\bibitem{mueller_water_2016}
N.~Mueller, A.~Lewis, D.~Roberts, S.~Ring, R.~Melrose, J.~Sixsmith,
  L.~Lymburner, A.~McIntyre, P.~Tan, S.~Curnow, and A.~Ip.
\newblock Water observations from space: {Mapping} surface water from 25years
  of {Landsat} imagery across {Australia}.
\newblock {\em Remote Sensing of Environment}, 174:341--352, March 2016.

\bibitem{pekel_high-resolution_2016}
Jean-François Pekel, Andrew Cottam, Noel Gorelick, and Alan~S. Belward.
\newblock High-resolution mapping of global surface water and its long-term
  changes.
\newblock {\em Nature}, 540(7633):418--422, December 2016.

\bibitem{isikdogan_rivamap_2017}
Furkan Isikdogan, Alan Bovik, and Paola Passalacqua.
\newblock {RivaMap}: {An} automated river analysis and mapping engine.
\newblock {\em Remote Sensing of Environment}, 202:88--97, December 2017.

\bibitem{yang_monthly_2020}
Xiucheng Yang, Qiming Qin, Hervé Yésou, Thomas Ledauphin, Mathieu Koehl,
  Pierre Grussenmeyer, and Zhe Zhu.
\newblock Monthly estimation of the surface water extent in {France} at a 10-m
  resolution using {Sentinel}-2 data.
\newblock {\em Remote Sensing of Environment}, 244:111803, July 2020.

\bibitem{ronneberger2015u}
Olaf Ronneberger, Philipp Fischer, and Thomas Brox.
\newblock U-net: Convolutional networks for biomedical image segmentation.
\newblock In {\em International Conference on Medical image computing and
  computer-assisted intervention}, pages 234--241. Springer, 2015.

\bibitem{randall2019geographic}
Mark Randall, Rasmus Fensholt, Yongyong Zhang, and Marina Bergen~Jensen.
\newblock {Geographic object based image analysis of WorldView-3 imagery for
  urban hydrologic modelling at the catchment scale}.
\newblock {\em Water}, 11(6):1133, 2019.

\bibitem{feng2018water}
Wenqing Feng, Haigang Sui, Weiming Huang, Chuan Xu, and Kaiqiang An.
\newblock Water body extraction from very high-resolution remote sensing
  imagery using deep u-net and a superpixel-based conditional random field
  model.
\newblock {\em IEEE Geoscience and Remote Sensing Letters}, 16(4):618--622,
  2018.

\bibitem{allen2015patterns}
George~H Allen and Tamlin~M Pavelsky.
\newblock Patterns of river width and surface area revealed by the
  satellite-derived north american river width data set.
\newblock {\em Geophysical Research Letters}, 42(2):395--402, 2015.

\bibitem{frasson2019global}
Renato Prata de~Moraes Frasson, Tamlin~M Pavelsky, Mark~A Fonstad, Michael~T
  Durand, George~H Allen, Guy Schumann, Christine Lion, R~Edward Beighley, and
  Xiao Yang.
\newblock Global relationships between river width, slope, catchment area,
  meander wavelength, sinuosity, and discharge.
\newblock {\em Geophysical Research Letters}, 46(6):3252--3262, 2019.

\bibitem{zhang2018road}
Zhengxin Zhang, Qingjie Liu, and Yunhong Wang.
\newblock Road extraction by deep residual u-net.
\newblock {\em IEEE Geoscience and Remote Sensing Letters}, 15(5):749--753,
  2018.

\bibitem{wei2019multi}
Sisi Wei, Hong Zhang, Chao Wang, Yuanyuan Wang, and Lu~Xu.
\newblock Multi-temporal sar data large-scale crop mapping based on u-net
  model.
\newblock {\em Remote Sensing}, 11(1):68, 2019.

\bibitem{yao2018pixel}
Wei Yao, Zhigang Zeng, Cheng Lian, and Huiming Tang.
\newblock Pixel-wise regression using u-net and its application on
  pansharpening.
\newblock {\em Neurocomputing}, 312:364--371, 2018.

\bibitem{wang2020weakly}
Sherrie Wang, William Chen, Sang~Michael Xie, George Azzari, and David~B
  Lobell.
\newblock Weakly supervised deep learning for segmentation of remote sensing
  imagery.
\newblock {\em Remote Sensing}, 12(2):207, 2020.

\bibitem{houborg2018cubesat}
Rasmus Houborg and Matthew~F McCabe.
\newblock A cubesat enabled spatio-temporal enhancement method (cestem)
  utilizing planet, landsat and modis data.
\newblock {\em Remote Sensing of Environment}, 209:211--226, 2018.

\bibitem{taudem_2015}
Ahmet~Artu Yıldırım, Dan Watson, David Tarboton, and Robert~M. Wallace.
\newblock A virtual tile approach to raster-based calculations of large digital
  elevation models in a shared-memory system.
\newblock {\em Computers \& Geosciences}, 82:78--88, September 2015.

\bibitem{McFeeters1996}
S.~K. McFeeters.
\newblock {The use of the Normalized Difference Water Index (NDWI) in the
  delineation of open water features}.
\newblock {\em International Journal of Remote Sensing}, 17(7):1425--1432,
  1996.

\bibitem{guizar-sicairos_efficient_2008}
Manuel Guizar-Sicairos, Samuel~T. Thurman, and James~R. Fienup.
\newblock Efficient subpixel image registration algorithms.
\newblock {\em Optics Letters}, 33(2):156--158, January 2008.

\bibitem{valada2016deep}
Abhinav Valada, Gabriel~L Oliveira, Thomas Brox, and Wolfram Burgard.
\newblock Deep multispectral semantic scene understanding of forested
  environments using multimodal fusion.
\newblock In {\em International Symposium on Experimental Robotics}, pages
  465--477. Springer, 2016.

\bibitem{yang2019deep}
Wenming Yang, Xuechen Zhang, Yapeng Tian, Wei Wang, Jing-Hao Xue, and Qingmin
  Liao.
\newblock Deep learning for single image super-resolution: A brief review.
\newblock {\em IEEE Transactions on Multimedia}, 21(12):3106--3121, 2019.

\bibitem{deudon2020highres}
Michel Deudon, Alfredo Kalaitzis, Israel Goytom, Md~Rifat Arefin, Zhichao Lin,
  Kris Sankaran, Vincent Michalski, Samira~E Kahou, Julien Cornebise, and
  Yoshua Bengio.
\newblock Highres-net: Recursive fusion for multi-frame super-resolution of
  satellite imagery.
\newblock {\em arXiv preprint arXiv:2002.06460}, 2020.

\end{thebibliography}

\clearpage

\clearpage

\appendix
\section{Appendix}
\subsection{LiDAR derived products}\label{sec:lidarder}
The derivative data layers derived from the 1 meter resolution 3DEP DEMs over all study areas are:
\begin{itemize}
\item Sink-filled DEM
\item Flow direction
\item Flow Accumulation
\item Synthetic reach network, thresholded to 10K square meters minimum catchment size
\item Horizontal distance buffer to reach network
\item Reach catchments
\item Strahler stream order
\item Height Above Ground (HAG).
\end{itemize}
Inputs to the WasserNetz models are the \ac{dem}, the Synthetic reach network, horizontal distance buffer to reach network and Height Above Ground (HAG).
\subsection{Co-registration}
\begin{figure*}[ht]
    \centering
    \includegraphics[width=\linewidth]{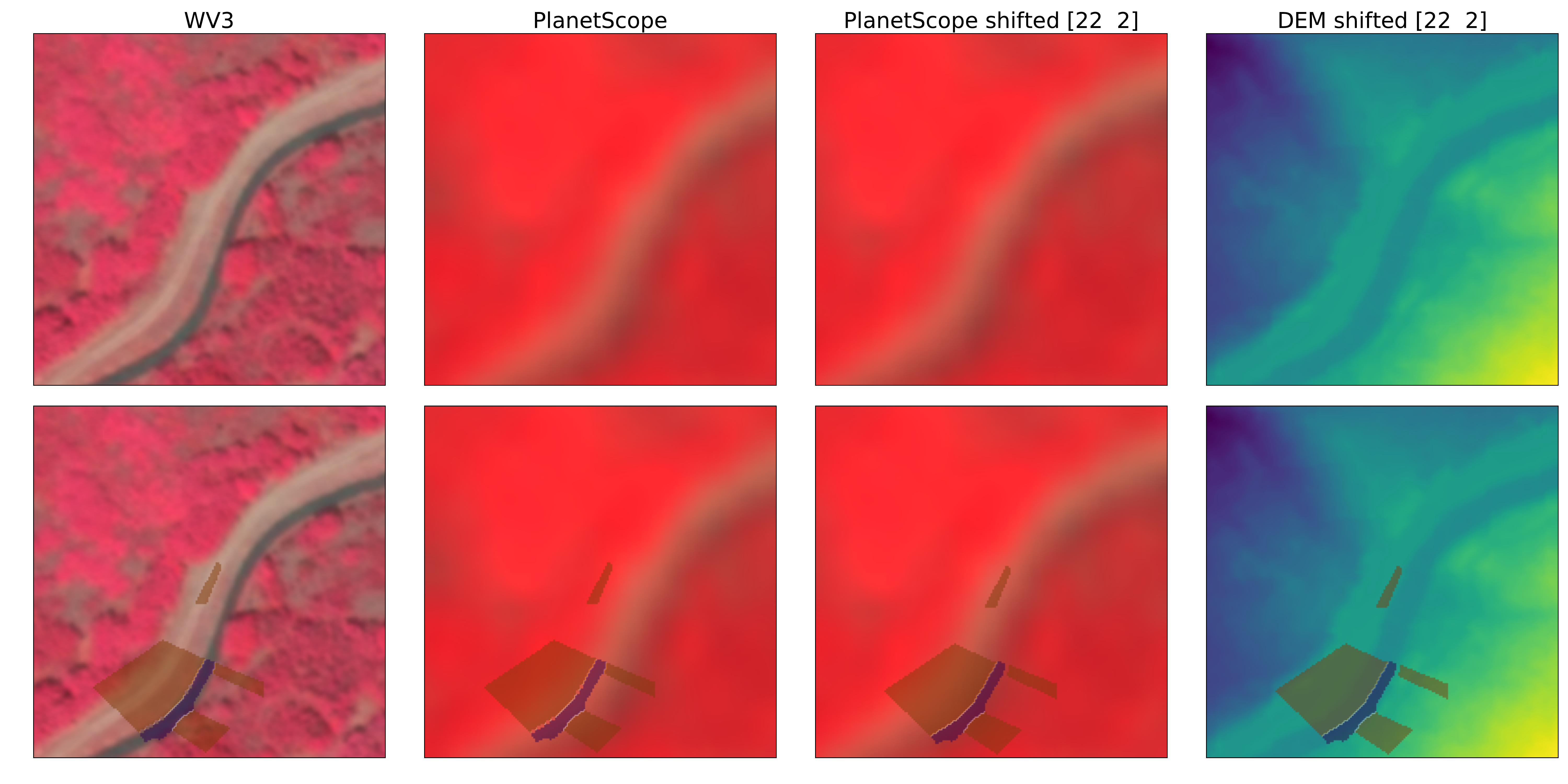}
    \caption{From left to right \ac{wv3} image, PlanetScope Image, PlanetScope image shifted to correct the polygon shift, \ac{dem} image shifted to correct the polygon shift. Second row shows the same image overlaid with training polygons.}
    \label{fig:correg}
\end{figure*}

Co-registration issues are observed due to the different orthorectification methods used by the different data sources. In particular, we observed shifts of WV images of around 1-20 meters compared with the DEM and the PlanetScope images. An example of these shifts is shown in Figure. \ref{fig:correg}. This misalignment will affect models with PlanetScope and/or \ac{dem} inputs trained with the polygons, which were annotated using WV images as a reference. To fix this issue, we use off-the-shelf image registration algorithms~\cite{guizar-sicairos_efficient_2008}, to shift the polygons for each example when training on Planet and/or \ac{lidar} derived data. Although with this method we fix large misalignments between Planet/DEM images and labeled polygons, small shifts might still be present as the different resolution of the sources limit the shift corrections.

\subsection{Precision Recall Curves}\label{sec:prcurves}

\begin{figure}[htbp]
    \centering
    \includegraphics[width=.45\linewidth]{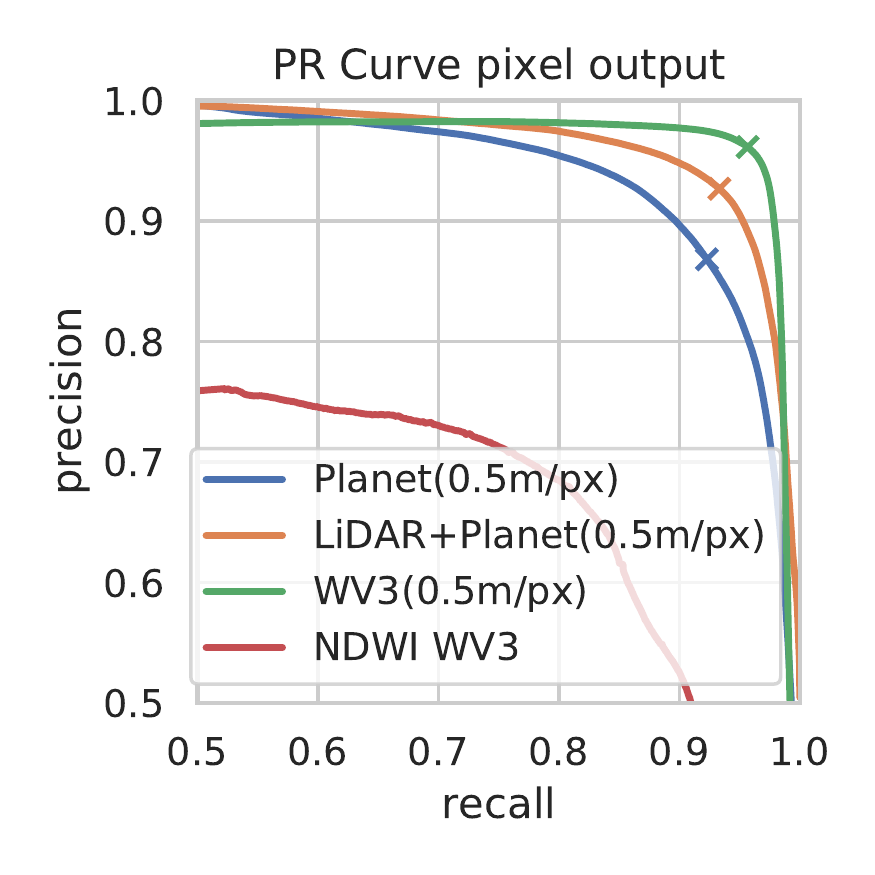}
    \includegraphics[width=.45\linewidth]{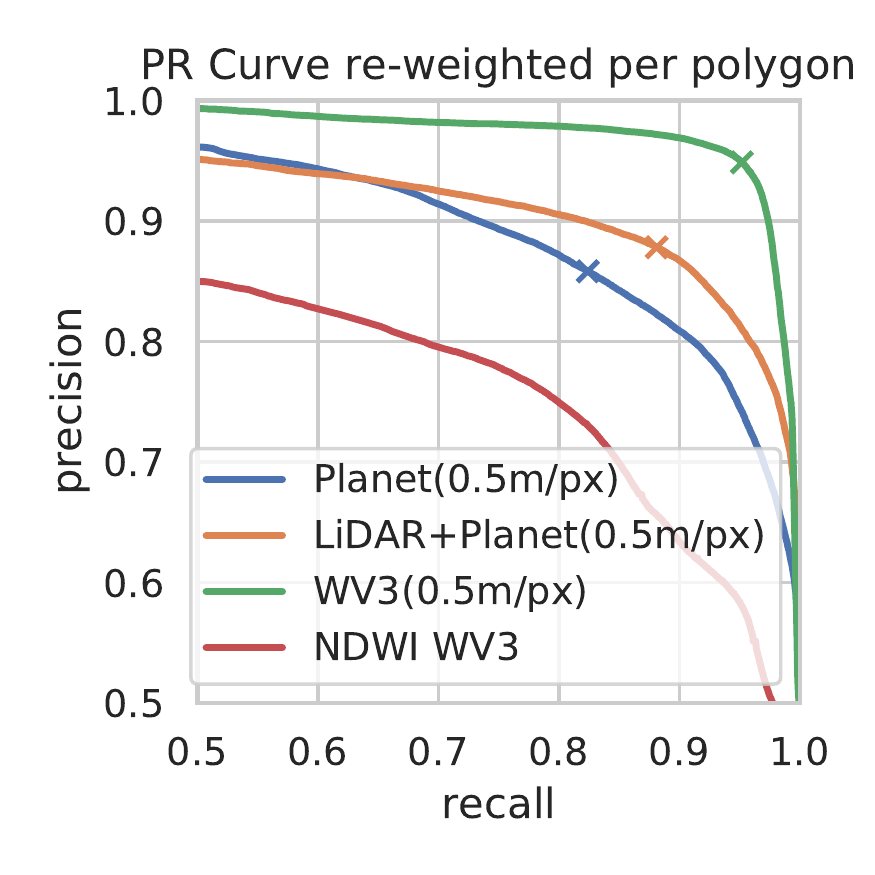}
    \caption{Precision-recall curves of the different models on the pixel-level output (left) and weighted per polygon, aka. polygon-level (right). Crosses shows the 0.5 threshold.}
    \label{fig:precrecall}
\end{figure}

A precision-recall curve shows the trade-off between precision and recall as we vary the threshold on the output of the network. Figure~\ref{fig:precrecall} shows precision recall on the raw pixel output and re-weighted by the number of pixels of each polygon. The crosses show the 0.5 probability threshold that was used in the models for simplicity. 

\subsection{WasserNetz}\label{sec:WasserNetz}

\begin{figure}[htbp]
  \centering
  \includegraphics[width=1\linewidth]{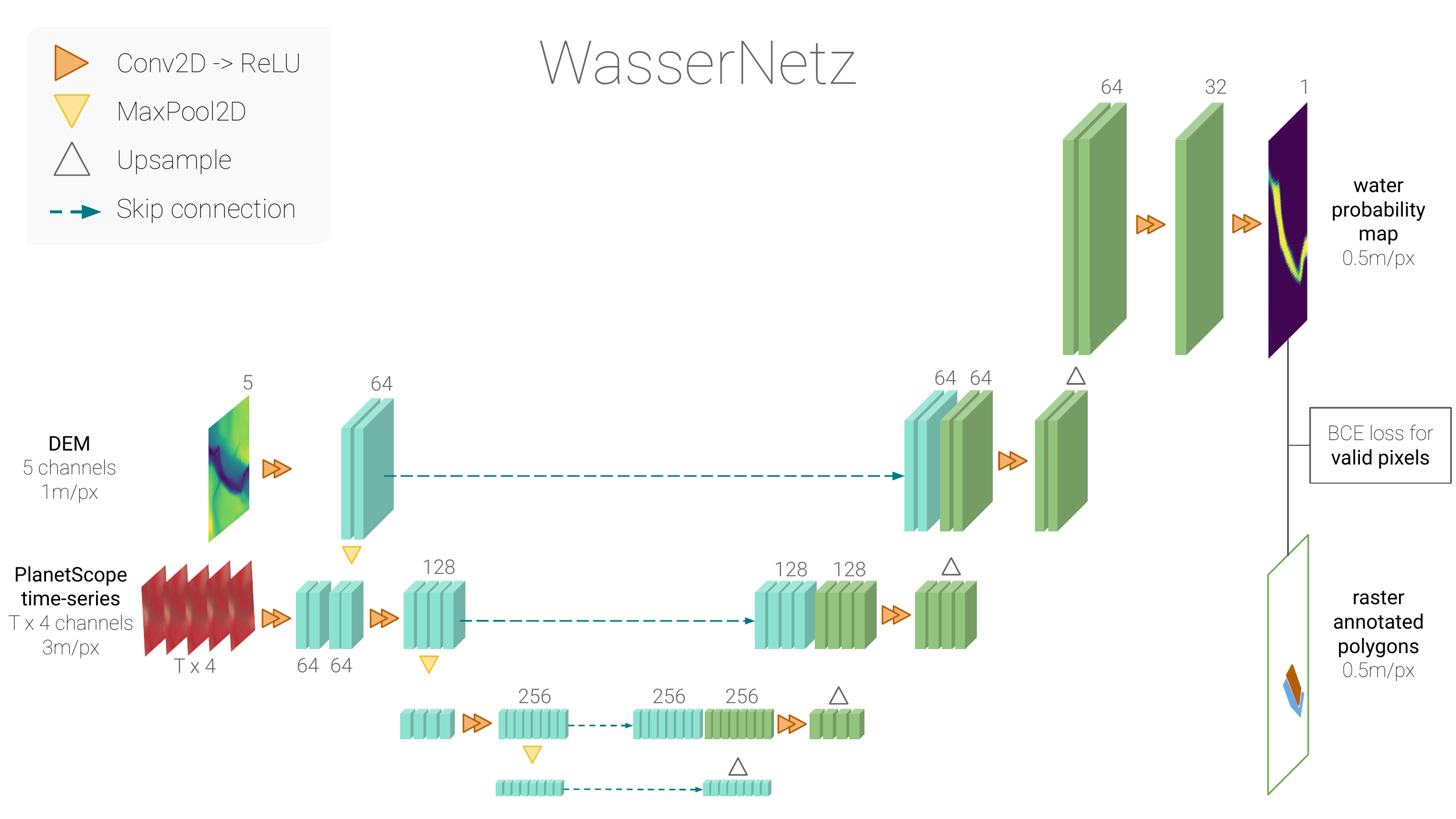}
  \caption{\small WasserNetz architecture.}
  \label{fig:WasserNetz}
\end{figure}

Fig~\ref{fig:WasserNetz} shows the architecture of WasserNetz, our variant of the U-Net architecture adapted for multi-sensor multi-resolution input fusion, and an output at some desired target resolution.

With a U-Net backbone, the design of WasserNetz follows common practices of image processing, such as resampling, which defines our multi-sensor multi-res input fusion~\cite{valada2016deep}.
We also resample the output to evaluate it at the native resolution (30-40cm/px) of our ground-truth -- a common practice in super-resolution \cite{yang2019deep,deudon2020highres}.

In light of our limited labeling resources, our sparse but pixel-based annotations allow a trade-off between full and weak supervision \cite{wang2020weakly}, see Section~\ref{sec:Data}.

\end{document}